# The Robust Gait of a Tilt-rotor and Its Application to Tracking Control -- Application of Two Color Map Theorem

Zhe Shen[1*] and Takeshi Tsuchiya[2]

[1,2] Department of Aeronautics and Astronautics, The University of Tokyo,
Tokyo, 113-8654, Japan (TEL: +81-080-3709-1024; E-mail: zheshen@g.ecc.u-tokyo.ac.jp)

**Abstract:** Ryll's tilt-rotor is a UAV with eight inputs; the four magnitudes of the thrusts as well as four tilting angles of the thrusts can be specified in need, e.g., based on a control rule. Despite of the success in simulation, conventional feedback linearization witnesses the over-intensive change in the inputs while applying to stabilize Ryll's tilt-rotor. Our previous research thus put the extra procedure named gait plan forward to suppress the unexpected changes in the tilting angles. Accompanying the Two Color Map Theorem, the tilting-angles are planned robustly and continuously. The designed gaits are robust to the change of the attitude. However, this is not a complete theory before further applying to the tracking simulation test. This paper further discusses some gaits following the Two Color Map Theorem and simulates a tracking problem for a tilt-rotor. A uniform circular moving reference is designed to be tracked by the tilt-rotor equipped with the designed robust gait and the feedback linearization controller. The gaits satisfying Two Color Map Theorem show the robustness. The results from the simulation show the success in tracking of the tilt-rotor.

**Keywords:** tilt-rotor, gait plan, two color map theorem, feedback linearization, simulation, tracking

## 1. INTRODUCTION

With the merit of the lateral thrusts, Ryll's tilt-rotor [1] attracted great attention in the past decade. While controlling this system utilizing feedback linearization, the tilting angles can change in unexpected over-intensive ways, which intrigues the birth of the gait plan of the tilt-rotor [2].

The gait plan is a procedure to pre-define the tilting angles (time-specified functions) of the tilt-rotor. Since there are eight inputs in the tilt-rotor, there will be only four magnitudes of the thrusts left to be specified by the controller after the gait plan procedure. A feedback linearization method is subsequently adopted to assign these four magnitudes of the thrusts.

Despite the fact that this control scenario successfully avoids the over-intensive change in the tilting angles from the theorem level, the subsequent feedback linearization method may encounter the singular decoupling matrix [3]–[5], causing the failure in the control result. The decoupling matrix is not invertible for several combinations of roll angle and pitch angle.

With this concern, a rule, Two Color Map Theorem [6], was put forward to help design the gait. Following Two Color Map Theorem, the designed gaits are guaranteed continuous and robust to the change of the attitude. The unacceptable combinations of the roll angles and pitch angles, which introduce the singular decoupling matrix, decrease, enabling the tilt-rotor to maneuver in wider attitude regions.

So far, no discussion on the tracking control has been addressed using the robust gaits, obeying Two Color Map Theorem. Tracking (dependently) the 6 degrees of freedom relying on the 4 inputs of the tilt-rotor can be realized with the help of the modified attitude-position decoupler put forward in our previous research [5]. Position X and Y are tracked relying on the coupling relationship with the degrees of freedom, roll, pitch, yaw, and altitude (Z), which are controlled independently.

In this paper, the robustness of several gaits following the Two Color Map Theorem is analyzed. One gait is then adopted to track a uniform circular moving reference. The designed gaits show their robustness to the attitude. The reference is successfully tracked in the simulation.

## 2. DYNAMICS OF THE TILT-ROTOR

Ryll's tilt-rotor is sketched in Fig. 1 [2].

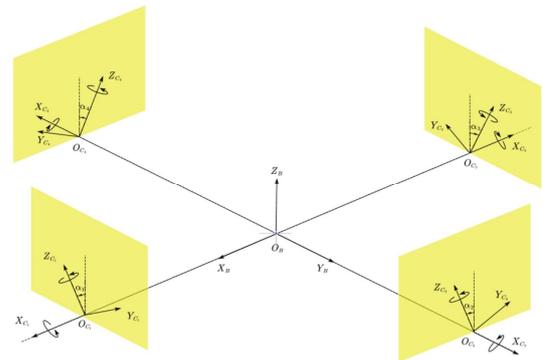

Fig. 1. Ryll's tilt-rotor.

The position of the tilt-rotor [3] is given by

$$\ddot{P} = \begin{bmatrix} 0 \\ 0 \\ -g \end{bmatrix} + \frac{1}{m} \cdot {}^W R \cdot F(\alpha) \cdot \begin{bmatrix} \varpi_1 \cdot |\varpi_1| \\ \varpi_2 \cdot |\varpi_2| \\ \varpi_3 \cdot |\varpi_3| \\ \varpi_4 \cdot |\varpi_4| \end{bmatrix}$$

$$\triangleq \begin{bmatrix} 0 \\ 0 \\ -g \end{bmatrix} + \frac{1}{m} \cdot {}^W R \cdot F(\alpha) \cdot w, \qquad (1)$$

where $P = [X \ Y \ Z]^T$ represents the position with respect to the earth frame, $m$ represents the total mass, $g$ represents the gravitational acceleration, $\varpi_i$, ($i = 1,2,3,4$) represents the angular velocity of the propeller $w = [\varpi_1 \cdot |\varpi_1|, \varpi_2 \cdot |\varpi_2|, \varpi_3 \cdot |\varpi_3|, \varpi_4 \cdot |\varpi_4|]^T$, and the term $^W R$ represents the rotational matrix.

The tilting angles $\alpha = [\alpha_1 \ \alpha_2 \ \alpha_3 \ \alpha_4]$. $F(\alpha)$ is given by

$$F(\alpha) = \begin{bmatrix} 0 & K_f \cdot s2 & 0 & -K_f \cdot s4 \\ K_f \cdot s1 & 0 & -K_f \cdot s3 & 0 \\ -K_f \cdot c1 & K_f \cdot c2 & -K_f \cdot c3 & K_f \cdot c4 \end{bmatrix}, \quad (2)$$

where $si = \sin(\alpha_i)$, $ci = \cos(\alpha_i)$, and ($i = 1,2,3,4$). $K_f$ ($8.048 \times 10^{-6} N \cdot s^2/rad^2$) is the coefficient of the thrust.

The angular velocity of the body [3] with respect to $\mathcal{F}_B$, $\omega_B = [p \ q \ r]^T$, is given by

$$\dot{\omega}_B = I_B^{-1} \cdot \tau(\alpha) \cdot w, \quad (3)$$

where $I_B$ is the matrix of moments of inertia, $K_m$ ($2.423 \times 10^{-7} N \cdot m \cdot s^2/rad^2$) is the coefficient of the drag, and $L$ is the length of the arm,

$\tau(\alpha)$
$$= \begin{bmatrix} 0 & L \cdot K_f \cdot c2 - K_m \cdot s2 & 0 & -L \cdot K_f \cdot c4 + K_m \cdot s4 \\ L \cdot K_f \cdot c1 + K_m \cdot s1 & 0 & -L \cdot K_f \cdot c3 - K_m \cdot s3 & 0 \\ L \cdot K_f \cdot s1 - K_m \cdot c1 & -L \cdot K_f \cdot s2 - K_m \cdot c2 & L \cdot K_f \cdot s3 - K_m \cdot c3 & -L \cdot K_f \cdot s4 - K_m \cdot c4 \end{bmatrix}$$
(4)

We refer the details of the dynamics and the feedback linearization control method to our previous research [3].

In our control scenario, the tilting angles, $\alpha$, are defined in advance in a separated procedure called gait plan [2]. The tilt-rotor is controlled using the four magnitudes of the thrusts by feedback linearization. Note that the decoupling matrix is singular for some attitude. The gait plan is supposed to generates a robust gait, which enlarges the acceptable attitude region.

## 3. GAIT PLAN

The gait of a tilt-rotor is a time-specified tilting angles, $\alpha_1(t)$, $\alpha_2(t)$, $\alpha_3(t)$, and $\alpha_4(t)$. Our previous research [6] puts forward a theorem, Two Color Map Theorem, to design robust gaits. The robust gait is a gait that has a large region of the acceptable attitudes, which introduce the invertible decoupling matrix.

### 3.1 Two Color Map Theorem

Firstly, design $\alpha_1(t)$ and $\alpha_2(t)$ on the $\alpha_1 - \alpha_2$ plane. For example, if we expect a periodic gait, then $(\alpha_1(t), \alpha_2(t))$ is an enclosed rectangular (defined with direction).

For a given time point at $t_1$, $(\alpha_1(t_1), \alpha_2(t_1))$ is subsequently determined. Then, $(\alpha_3(t_1), \alpha_4(t_1))$ is to be determined to finish designing a gait.

It is proved [6] that there are two $(\alpha_3(t_1), \alpha_4(t_1))$ corresponding to a defined $(\alpha_1(t_1), \alpha_2(t_1))$, in general, to design a robust gait.

These two corresponding $(\alpha_3, \alpha_4)$, corresponding to $(\alpha_1, \alpha_2)$ on the entire $\alpha_1 - \alpha_2$ diagram, can be classified into two groups, 'red $(\alpha_3, \alpha_4)$' and 'blue $(\alpha_3, \alpha_4)$'.

Interestingly, $\alpha_3$ in the same color, all the red $\alpha_3$ or all the blue $\alpha_3$, lies on the same plane in the $O\alpha_1\alpha_2\alpha_3$ coordinate system. Similarly, $\alpha_4$ in the same color, all the red $\alpha_4$ or all the blue $\alpha_4$, lies on the same plane in the $O\alpha_1\alpha_2\alpha_4$ coordinate system.

Obviously, to design a robust gait, the adjacent $(\alpha_3, \alpha_4)$ must be in the same color for the interested $(\alpha_1, \alpha_2)$. This is the simplified Two Color Map Theorem without crossover. Note that the cases considering crossover is also considered in our previous research [6], which is beyond the scope of this paper. Fig. 2 and Fig. 3 display the entire map for the first time.

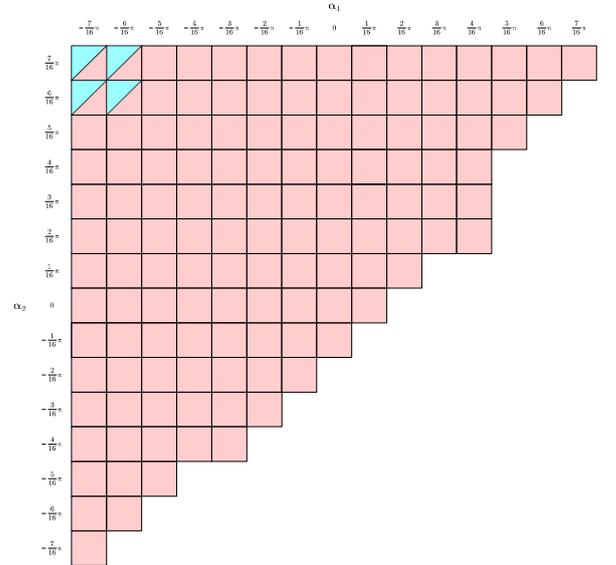

Fig. 2. Two Color Map with positive residual.

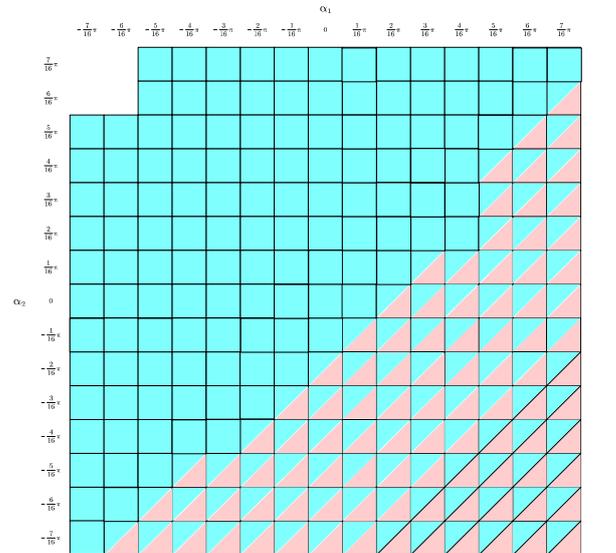

Fig. 3. Two Color Map with negative residual.

## 3.2 Robustness analysis

In this research, the robustness of four gaits is analyzed. They are Gait 1, Gait 2, and Gait 3, illustrated in Fig. 4, where $(\alpha_3, \alpha_4)$ of Gait 1 is blue $(\alpha_3, \alpha_4)$, $(\alpha_3, \alpha_4)$ of Gait 2 and Gait 3 are red $(\alpha_3, \alpha_4)$.

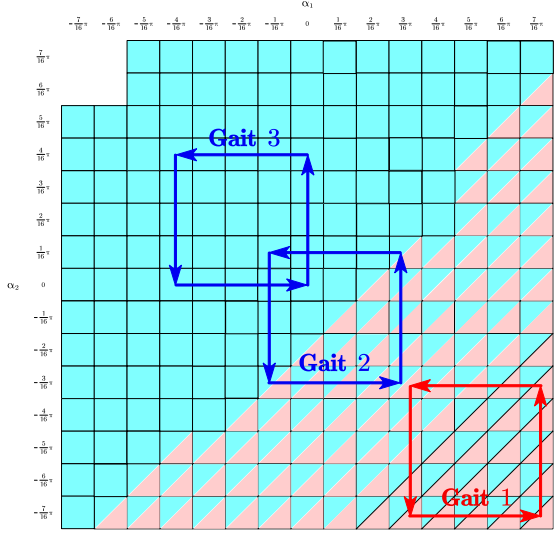

Fig. 4. Three gaits analyzed in this paper. $(\alpha_3, \alpha_4)$ of Gait 1 is blue $(\alpha_3, \alpha_4)$. $(\alpha_3, \alpha_4)$ of Gait 2 and Gait 3 are red $(\alpha_3, \alpha_4)$.

Figs. 5 ~ 7 plot the unacceptable attitudes (red curves) for Gait 1 ~ Gait 3, respectively. The unacceptable attitudes will introduce the singular decoupling matrix in feedback linearization, which should be prohibited. The tilt-rotor can only maneuver in the attitude region, which is not occupied by these attitude curves.

In comparison, we create the biased gait for each gait; the biased gait is generated by scaling $\alpha_3$ and $\alpha_4$ to their 80%. The unacceptable attitudes (blue curves) for the biased Gait 1 ~ biased Gait 3 are also displayed in Figs 5 ~ 7, respectively.

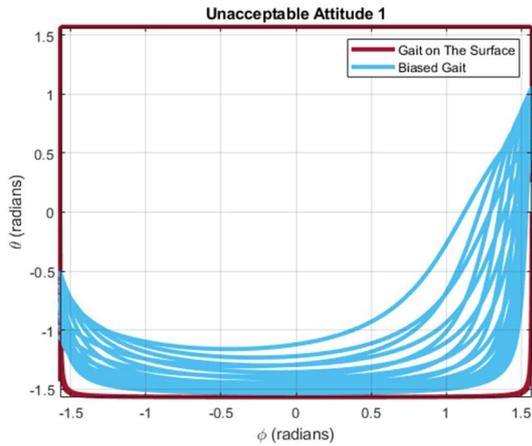

Fig. 5. The red curves represent the unacceptable attitude of Gait 1. The blue curves represent the unacceptable attitude of the biased Gait 1.

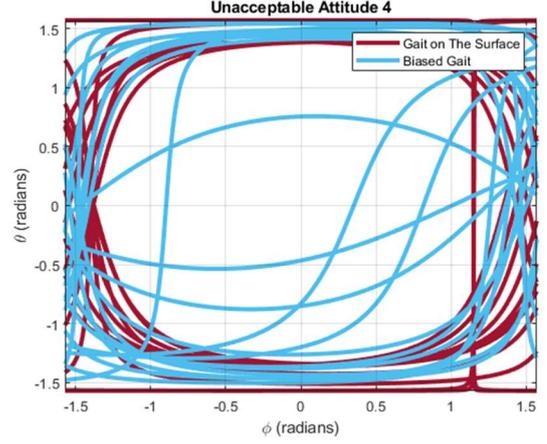

Fig. 6. The red curves represent the unacceptable attitude of Gait 2. The blue curves represent the unacceptable attitude of the biased Gait 2.

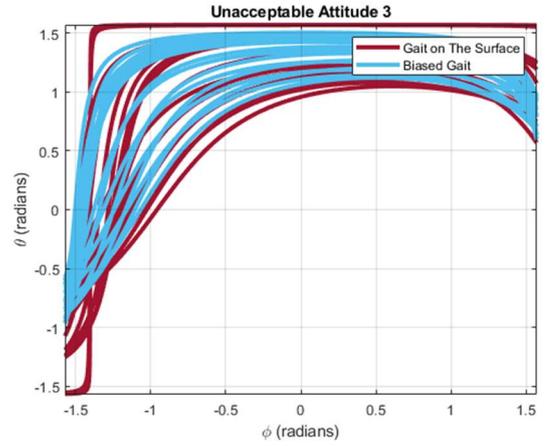

Fig. 7. The red curves represent the unacceptable attitude of Gait 3. The blue curves represent the unacceptable attitude of the biased Gait 3.

In general, the acceptable attitude region enlarges if the gait following the Two Color Map Theorem, especially for the gaits with the red $(\alpha_3, \alpha_4)$. The decoupling matrix introduced by the biased Gait 2 is very sensitive to the attitude. The tilt-rotor can be less likely to be stabilized by feedback linearization for these two biased gaits.

## 4. TRACKING SIMULATION

### 4.1 Reference and feasible gait

The reference set in the simulation is a uniform circular moving reference, the position of which is define in Eqs. (5) ~ (7).

$$x_r = 5 \cdot \cos(0.1 \cdot t). \quad (5)$$
$$y_r = 5 \cdot \sin(0.1 \cdot t). \quad (6)$$
$$z_r = 0. \quad (7)$$

The velocities of the reference along each dimension are received by calculating the derivatives of the velocities. Note that the acceleration of the reference is set as zero in the simulation test. We refer the detail of the controller as well as the modified attitude-position decoupler to our previous research [3], [5].

Besides, although we planned 3 gaits, Gait 1, Gait 2, and Gait 3. The result shows that only Gait 1 works in tracking. The rest gaits receive singular decoupling matrix, which is caused by the saturation of the input (non-zero saturation). Further discussions on the saturations of the input and the stability of feedback linearization can be referred to [7].

The initial position of the tilt-rotor is at $(x_i, y_i, z_i) = (0,0,0)$. The initial angular velocities of the propellers are insufficient to compensate the gravity. The tilt-rotor is expected to stabilize its altitude as well as to track the reference.

### 4.2 Tracking simulation result

Fig. 8 displays the gait that we adopted (Gait 1) in one period. And, the angular velocity history of each propeller is in Fig. 9.

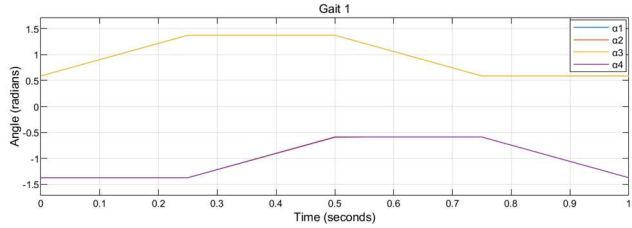

Fig. 8. Gait 1 within the first period.

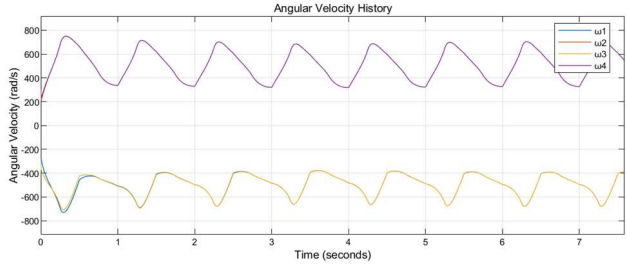

Fig. 9. Angular velocity history.

The trajectory (red curve) of the tilt-rotor in the simulation is illustrated in Fig. 10, where the reference is represented by the blue curve.

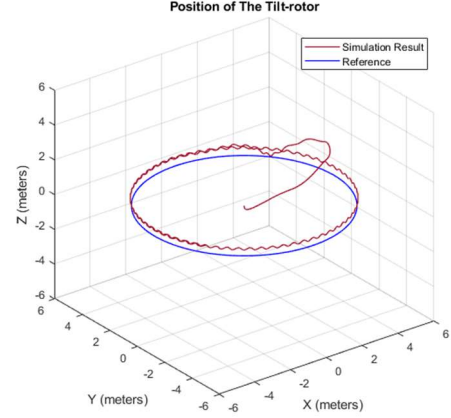

Fig. 10. Tracking result of the tilt-rotor in the simulation. The blue circle is the reference. The red curve represents the actual trajectory of the tilt-rotor while tracking.

The dynamic state error is presented in Fig. 11. It is close to zero after sufficient time, indicating that the tilt-rotor approaches the reference.

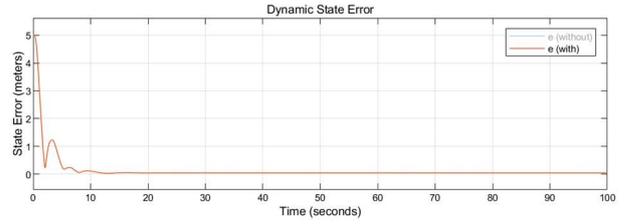

Fig. 11. The dynamic state error of the tilt-rotor. It is defined by the difference between the reference and the actual position of the tilt-rotor.

## 5. CONCLUSION

In this paper, we further verified the effectiveness of the Two Color Map theorem. The whole map is displayed for the first time. The robustness of several robust gaits are further compared with the ones not on the Map. A tracking simulation is made for the first time.

In general, the gaits following Two Color Map Theorem receive higher robustness to the attitude. While some of these gaits still receive singular decoupling matrix in feedback linearization because of the activation of the input saturation.

Although the initial condition is unstable for the tilt-rotor. It successfully compensates the gravity and tracks the uniform circular moving reference adopting the robust gait, which follows the Two Color Map Theorem. The dynamic state error decreases near to zero after sufficient time.

The angular velocities of the propeller also change periodically according to the gait. All the inputs are continuous.